\documentclass[sigconf]{acmart}

\renewcommand\footnotetextcopyrightpermission[1]{} 

\AtBeginDocument{%
  }

\setcopyright{acmlicensed}
\copyrightyear{2024}
\acmYear{2024}
\acmDOI{XXXXXXX.XXXXXXX}

\acmSubmissionID{123-A56-BU3}

\usepackage{float}
\begin{document}

\title{CACA Agent: Capability Collaboration based AI Agent}


\author{Peng Xu}
\affiliation{%
  \institution{State Key Lab of Networking and Switching Technology}
  \institution{Beijing University of Posts and Telecommunications}
  \city{Beijing}
  \country{China}}
\email{xupeng@bupt.edu.cn}

\author{Haoran Wang}
\authornote{Corresponding author.}
\affiliation{%
  \institution{State Key Lab of Networking and Switching Technology}
  \institution{Beijing University of Posts and Telecommunications}
  \city{Beijing}
  \country{China}
}
\email{wanghr@bupt.edu.cn}

\author{Chuang Wang}
\affiliation{%
  \institution{State Key Lab of Networking and Switching Technology}
  \institution{Beijing University of Posts and Telecommunications}
  \city{Beijing}
  \country{China}}
\email{wangchuang0418@outlook.com}

\author{Xu Liu}
\affiliation{%
  \institution{China Academy of Industrial Internet}
  \city{Beijing}
  \country{China}}
\email{liuxu@china-aii.com}


\begin{abstract}
As AI Agents based on Large Language Models (LLMs) have shown potential in practical applications across various fields, how to quickly deploy an AI agent and how to conveniently expand the application scenario of AI agents has become a challenge. Previous studies mainly focused on implementing all the reasoning capabilities of AI agents within a single LLM, which often makes the model more complex and also reduces the extensibility of AI agent functionality. In this paper, we propose CACA Agent (Capability Collaboration based AI Agent), using an open architecture inspired by service computing. CACA Agent integrates a set of collaborative capabilities to implement AI Agents, not only reducing the dependence on a single LLM, but also enhancing the extensibility of both the planning abilities and the tools available to AI agents. Utilizing the proposed system, we present a demo to illustrate the operation and the application scenario extension of CACA Agent.

\end{abstract}




\maketitle
\section{Introduction}
In recent years, Large Language Models (LLMs) have demonstrated remarkable learning and reasoning capabilities, achieving significant accomplishments. Consequently, researchers have implemented decision-making functions in AI Agents based on these models. By enhancing memory and tool functionalities, AI Agents are now capable of representing humans to complete various tasks. 
AI Agents based on LLMs have shown potential in practical applications across various fields.

The planning abilities of AI Agents are key factors affecting the quality of task handling. In some studies, AI Agents efficiently handle complex tasks by setting sub-goals and using decomposition strategies\cite{2022Chain}. They are capable of self-reflection on past interactions with the environment, improving the quality of future decisions and actions\cite{huang2022inner}\cite{yao2022react}. However, a significant drawback of this approach is its failure to equip AI Agents with a priori factual information. This often results in unrealistic outputs, attributable to the hallucinatory tendencies of large models.

AI Agents also need to interact with the external environment through various tools. LLMs have shown great potential in tool usage capabilities\cite{patil2023gorilla}\cite{qin2023toolllm}
\cite{schick2023toolformer}\cite{song2023restgpt}\cite{wang2023survey}.
AI Agents can greatly expand their application scenarios by calling open APIs of other applications, such as using weather APIs to obtain real-time weather data and using web search APIs to access internet information, providing users with more personalized and comprehensive services. However, enabling AI Agents to quickly master the use of new tools is challenging, often requiring a significant amount of new training data and model training to achieve good results.

To mitigate these challenges, we have developed a system known as the Capability Collaborative AI Agent System (CACA Agent). In this system, a series of capabilities work together to provide AI Agent functionalities to users. Firstly, we introduced Planning Capability and Methodology Capability. Planning Capability offers planning abilities based on LLMs, while Methodology Capability focuses on providing Planning Capability with factual information related to planning and the ability to interact with experts. Secondly, we have drawn on the service computing architecture by introducing Tools Capability, Tool Broker, and Tool Service into the "Registration-Discovery-Invocation" framework, thus supporting dynamically expand tools of AI Agents. \href{https://github.com/free4inno/CACA-Agent}{Source codes of CACA Agent can be found here.}\footnote{\url{https://github.com/free4inno/CACA-Agent}}

\section{Related Works}

Planning capability refers to the ability to decompose complex tasks into simpler subtasks. Some studies, such as Chain of Thought \cite{2022Chain}, utilize the reasoning abilities of LLMs, inspiring the models to plan and act through prompting words. However, this approach can result in 'hallucinations,' where the generated content conflicts with factual information. Methods to solve this problem include enabling models to adjust plans based on external environments (as in ReAct\cite{yao2022react}) or human feedback (such as Inner Monologue\cite{huang2022inner}). Nevertheless, feedback-based planning strategies are primarily suitable for simple tasks. For complex tasks, large models often need to refer to experts to find solutions.

To address these problems, we designed methodology capability to enhance planning capability. Methodology Capability enables the rapid expansion of the knowledge base concerning planning processes and provides interfaces for expert feedback, facilitating the convenient introduction of new process knowledge.

The ability to utilize tools is crucial for the functional expansion of large models and their effective interaction with the real world. Studies have demonstrated that LLMs are capable of autonomously learning tool usage through APIs (such as ToolFormer\cite{schick2023toolformer}), select tools from API documentation (as exemplified by  Gorilla\cite{patil2023gorilla}), or control actual applications by connecting to RESTful APIs (like RestGPT\cite{song2023restgpt}). ToolLLM\cite{qin2023toolllm} provides a comprehensive tool utilization framework that includes data construction, model training, and evaluation. However, these methods have limitations, such as a finite definition of tools, datasets primarily targeted at machine learning APIs, the need for data cleaning and annotation to expand tools, and an inability to meet more complex and diverse needs.

To address these issues, we drew inspiration from the concept of service computing and designed a tool capability framework based on the "Registration-Discovery-Invocation" Mechanism. This framework facilitates rapid expansion of tool services and leverages the inferential capabilities of LLMs for tool selection. This approach not only enhances the system's flexibility but also significantly simplifies the integration and addition of tools, making the process more efficient and convenient.

\section{System Architecture}

\subsection{Overall System Design}
The core concept of the CACA Agent system design centers around a collaborative set of capabilities aimed at implementing AI Agents. The CACA Agent system boasts several advantages:

\begin{enumerate}
  \item The functionality of an AI Agent is accomplished through the collaboration of a series of components, each possessing different capabilities. These components can be independently designed, implemented, deployed, and upgraded, and can be provided by different vendors.
  \item The Methodology Capability enhances the system by enabling rapid expansion of the process knowledge base, thereby improving planning quality with clearer process insights and facilitating the expansion of planning capabilities.
  \item Inspired by service computing, a tool capability framework based on the 
  "Registration-Discovery-Invocation" Mechanism is designed to provide flexibility and extensibility of tools.
\end{enumerate}

Our system design is illustrated in Figure 1. A set of collaborative capabilities is introduced to implement AI Agents\cite{weng2023prompt}.

\begin{figure}[h]
  \centering
  \includegraphics[width=\linewidth]{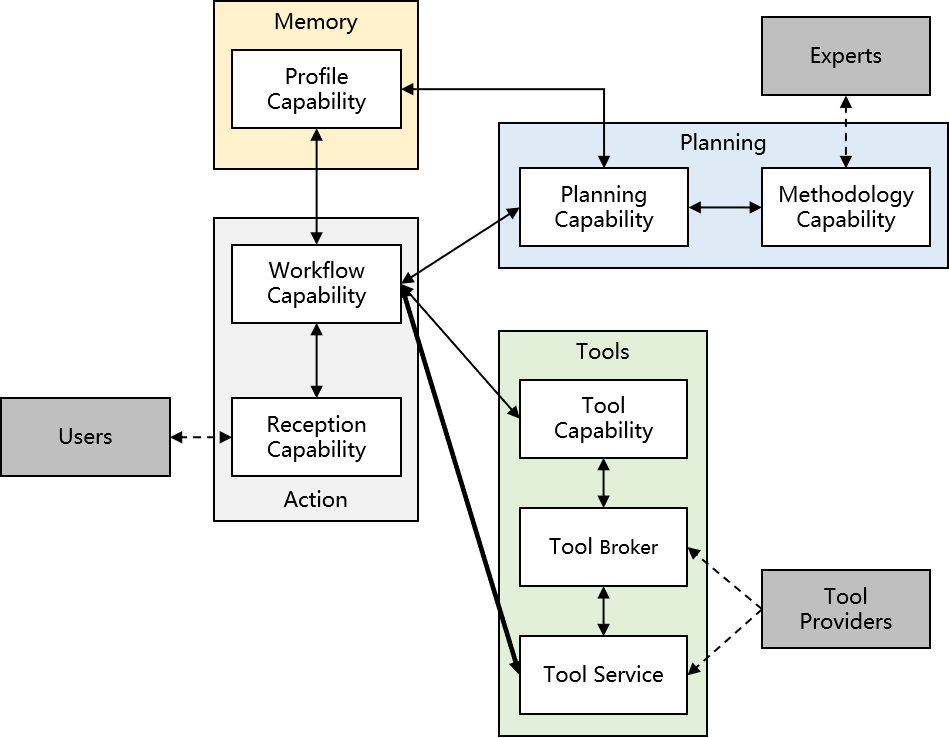}
  \caption{System Design Diagram}
  \Description{2}
\end{figure}

\textbf{[Action]} Workflow Capability and Reception Capability are introduced to fulfill the functions of Action in the AI Agent.

\textbf{Reception Capability} primarily provides a user interface, receives user requests, and responds accordingly. Importantly, Reception Capability does not directly handle the details of user requests but converts them into structured task descriptions upon receipt, then passes the task to Workflow Capability to create a workflow instance. Once Workflow Capability completes the relevant processing steps, Reception Capability feeds back the results of the task to the user, realizing an effective user request processing workflow.

\textbf{Workflow Capability} is mainly used to manage workflow instances for each user request. In the CACA Agent, Workflow instances use the same template. They first interact with Planning Capability to obtain specific processing procedures, then consult Tool Capability for relevant tool information according to the needs of each operation in the process, and finally complete interface calls based on this information, eventually responding to the users with the relevant processing results.

\textbf{[Planning]} Planning Capability and Methodology Capability work together to provide the functions of Planning in the AI Agent.


\textbf{Planning Capability} is primarily realized through LLMs, operating by providing a process flow for handling requests based on user queries. The Planning Capability relies on LLMs to implement the necessary reasoning abilities for planning. To ensure that the Planning Capability can deliver more accurate processing flows, and also to facilitate the convenient expansion of the AI Agent's application capabilities, it integrates domain knowledge obtained from the Methodology Capability. This integration involves breaking down complex tasks into a series of practical, feasible sub-steps, and then transmitting this task planning back to the Workflow Capability. The workflow process of the Planning Capability can be described by the following formula:


 \begin{align}
    PlanCap(Task,Methodology) \rightarrow [Proc(ST_1),\ldots,Proc(ST_n)]
\end{align}


To accommodate various conditions, choices, and repetitive execution needs, $Proc(ST_i)$ is further broken down into:

\begin{align}
 \begin{split}
    Proc(ST_i):(Execute(ST_i),Branch(ST_i), Loop(ST_i))
  \end{split}
\end{align}

Here, each component has a specific function. $Execute(ST_i)$ is responsible for directly executing each sub-task, ensuring each decomposed task unit is implemented. 
$Branch(ST_i)$ represents applying branching logic, choosing different execution paths based on different situations. $Loop(ST_i)$ allows for the repetition of certain sub-tasks under specific conditions, providing a looping mechanism for scenarios requiring repeated execution of the same operation, such as batch processing or iterative optimization.

\textbf{Methodology Capability} maintains knowledge about processing procedures for a variety of application requests across different scenarios.  Simultaneously, it provides a user interface to experts, allowing them to input knowledge regarding processing procedures for specific scenarios. We propose a data structure designed to enable experts to provide and update their professional knowledge.

This data structure comprises several key components: (1) \textbf{Description} provides detailed background of the application scenario. (2) \textbf{ProcessSteps} is an ordered list outlining each step with title, description, and required data.(3) \textbf{DecisionPoints} specifically identifies key decision points in the process along with their decision logic. (4) \textbf{Rules} section allows experts to input specific conditions that influence the process path. (5) \textbf{Exception} section describes methods for addressing potential exceptions in the process. (6) \textbf{Suggestions} provides a space for experts to offer process optimization suggestions. (7) \textbf{Reference} enables uploading relevant documents for a more comprehensive explanation of the process.


\textbf{[Memory]} Profile Capability provides the long-term memory for users. Methodology Capability possesses the knowledge of processes and Tool Capability saves the functions of various registered tools. As to the short-term memory about the processing of each user request, it is maintained by Workflow Capability.

\textbf{Profile Capability} in the CACA Agent system is tasked with managing the configuration information the system relies on. Its main responsibility is to maintain essential information, such as the basic configuration of the system and accounts required for operations. Distinguishing between long-term and short-term memory, the Profile Capability primarily offers long-term memory. Within the CACA Agent system, short-term memory is predominantly managed by the Workflow Capability within the workflow instances for each user request.

\textbf{[Tool]} The Tools function is facilitated through the collaborative efforts of the Tool Capability, Tool Broker, and Tool Service. Drawing inspiration from Service Computing, the CACA Agent system employs a "Registration-Discovery-Invocation" framework.

\textbf{Tool Capability} acts as both a tool registry and discovery center.   It handles tool registration requests from the Tool Broker, ensuring the upkeep and management of information related to various tools. Additionally, it manages tool discovery requests from Workflow Capability, providing relevant tool information based on specific operational needs. Tool Capability utilizes the reasoning abilities of LLMs to analyze task descriptions and information on registered tools, selecting the most appropriate tool and returning information such as the relevant invocation parameters. This process is encapsulated in the following formula:

\begin{align}
    \{T_{\text{selected}}, T_{\text{param}}[ \text{input} ]\} &= \text{DiscoverTool}( U_{\text{req}}, \{ T_{\text{registered}} \} ) 
\end{align}

where $T_{selected}$ represents the selected tool, $T_{param}$ denotes the parameters needed to invoke the tool, $U_{req}$ is the user's request, and $T_{registered}$ signifies the registered tools.

\textbf{Tool Service} specifically provides relevant processing capabilities, primarily through service APIs. 

\textbf{Tool Broker} acts as an intermediary for one or more Tool Services,  facilitating their registration with Tool Capability, and also ensuring the validity of the services it represents through mechanisms like heartbeats. It is noteworthy that tool registration by the Tool Broker is typically manually initiated by the tool providers. After deploying and configuring the relevant tools, providers fill in the tool information through the interface provided by Tool Broker and initiate a tool registration request.


\subsection{Key Workflows}

\begin{figure}[h]
  \centering
  \includegraphics[width=\linewidth]{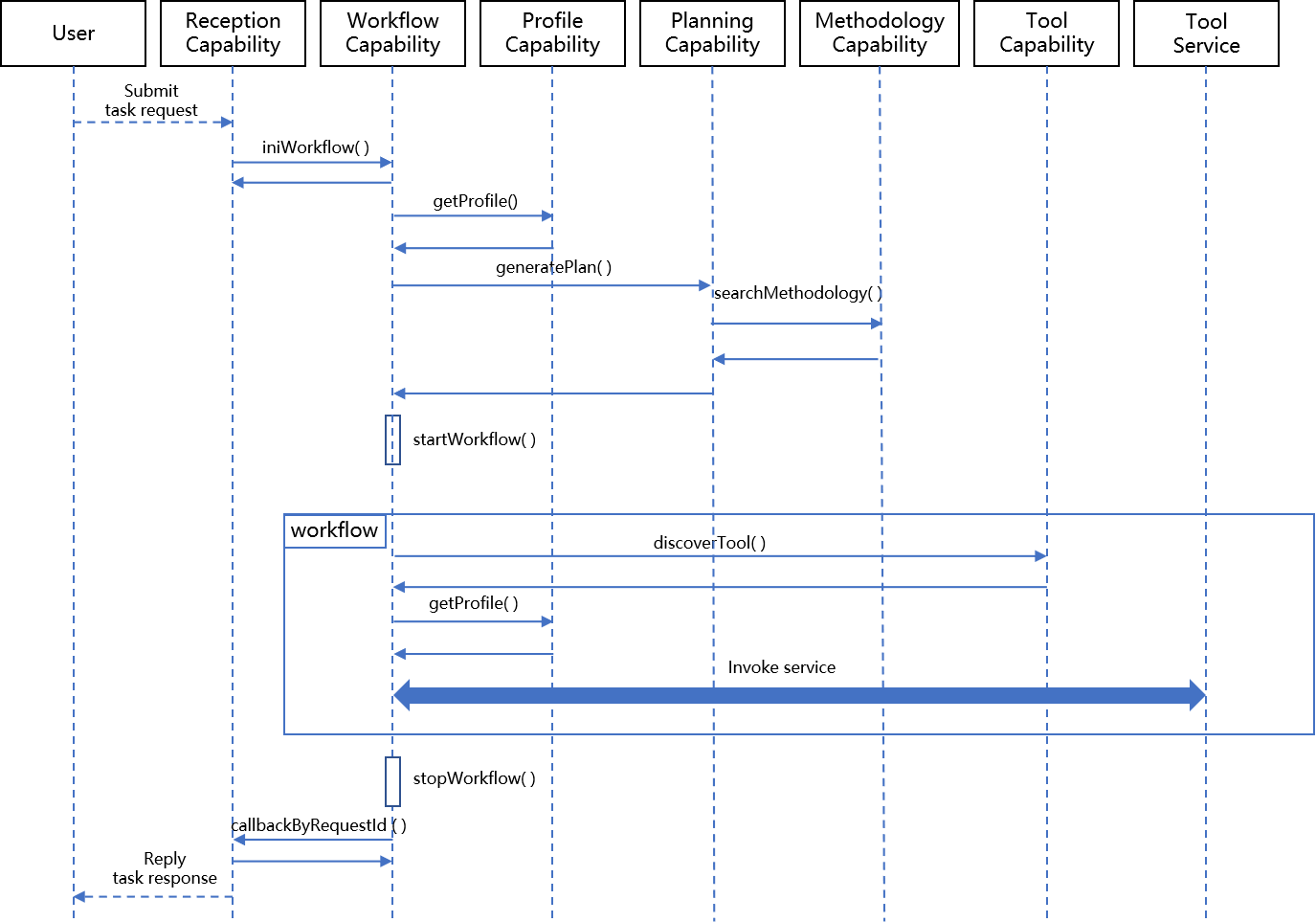}
  \caption{User Request Processing}
  \Description{2}
\end{figure}

The user request processing flow is illustrated in Figure 2. After a user submits a request, it is initially processed by the Reception Capability, and then handed off to the Workflow Capability. The Workflow Capability subsequently queries factual data from the Profile Capability, creates a new workflow instance, and consults the Planning Capability for task execution planning. The Planning Capability, by integrating domain knowledge, divides the task into sub-steps and provides feedback to the Workflow Capability.

Based on the task planning, the Workflow Capability initiates an asynchronously executed workflow. During the processing of sub-steps, it queries the Tool Capability for the most appropriate tools and parameters, generates tool invocation parameters, and then calls the tool service to complete the sub-steps. Upon completion of the workflow process, the Workflow Capability reports the execution results to the Reception Capability. Finally, the Reception Capability conveys the processed results back to the user.

\section{Demo}

\begin{figure}[h]
  \centering
  \includegraphics[width=\linewidth]{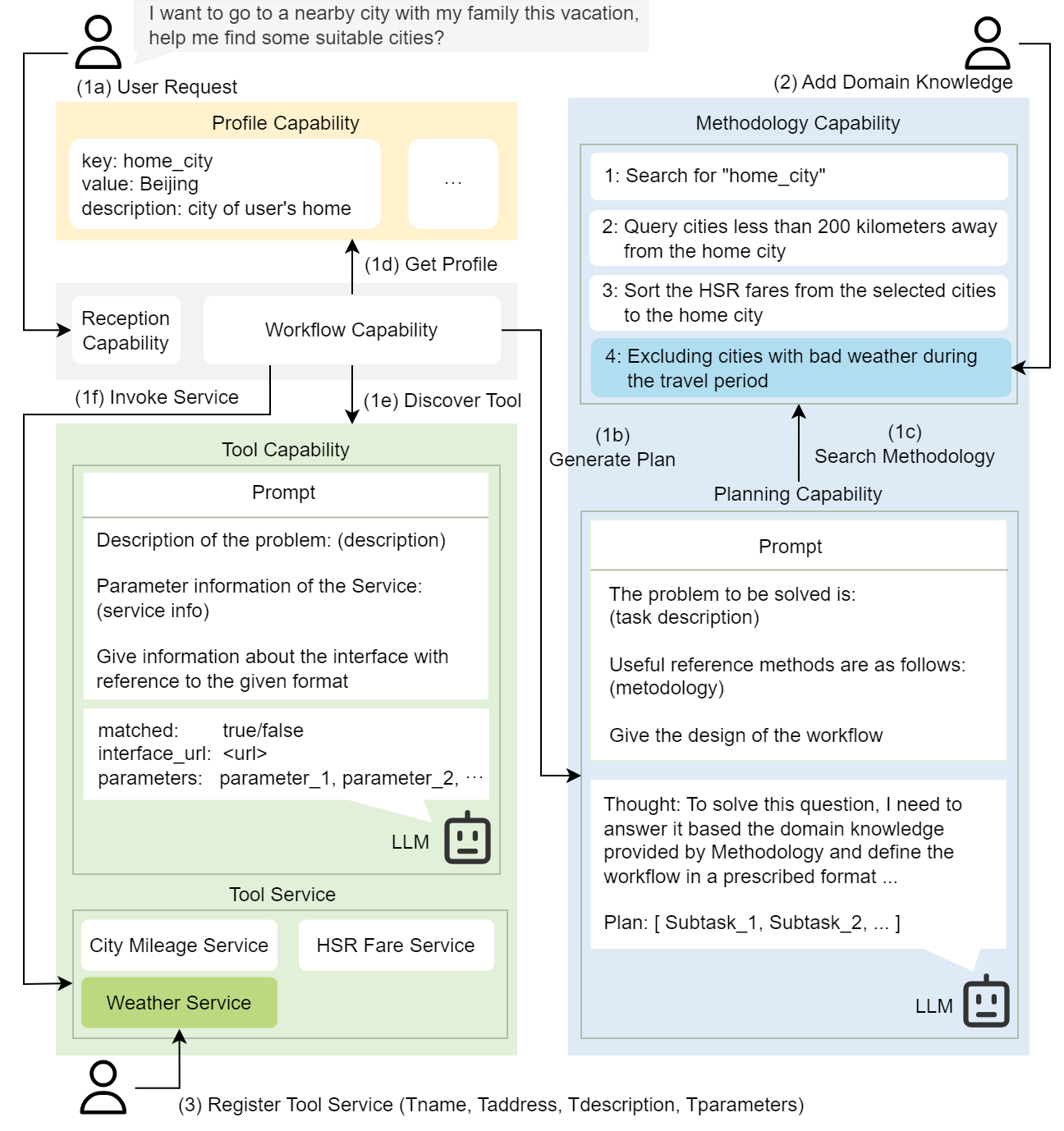}
  \caption{Workflow of CACA Agent}
  \Description{3}
\end{figure}

\subsection{Use Case}

We illustrate the workflow of the CACA Agent through a task requirement for recommending a destination for a family's short-distance travel. The demonstration process is depicted in Figure 3. The demonstration is divided into three scenarios: Scenario 1 showcases the workflow of the CACA Agent, highlighting the interactions between the planning capability and the LLM, as well as the interactions between the tool capability and the LLM. In Scenario 2, we demonstrate how the planning ability can be extended to consider weather conditions when recommending a destination. Scenario 3 showcases the extensibility of the tool.

\subsection{Scenario 1: Workflow of CACA Agent}
On the webpage provided by the Reception Capability, a user asks the CACA Agent: "I want to go to a nearby city with my family this vacation, can you help me find some suitable cities?" The Reception Capability forwards the request to the Workflow Capability for processing (Figure 3 (1a)). Upon receiving the user’s query, the Workflow Capability consults the Planning Capability to devise a plan to address the request (Figure 3 (1b)). The Planning Capability then consults the Methodology Capability for methods to solve the issue (Figure 3 (1c)). Once a method is identified, it is returned to the Planning Capability, which, using this information, generates a structured workflow description with the assistance of an LLM.

For the first subtask in the workflow, the Workflow Capability queries the Profile Capability for the user’s home address information (Figure 3 (1d)). For the second and third subtasks, the Workflow Capability sequentially queries the Tool Capability for available tool services (Figure 3 (1e)). At this point, several tools registered in the Tool Broker meet the subtask requirements. The Tool Capability then sends information about these tool services to the Workflow Capability. Having received this information, the Workflow Capability sequentially invokes these two tool services, thus completing the entire workflow process (Figure 3 (1f)).

\subsection{Scenario 2: Extensibility of Planning Ability}

It is important to note that the method process described above does not consider weather conditions during the trip. To comprehensively address the user’s needs, we add a subtask named "Excluding cities with adverse weather during the travel period." This addition is facilitated through the expert interaction page provided by the Methodology Capability (Figure 3 (2)).

\subsection{Scenario 3: Extensibility of Tools}
However, this new subtask lacks a corresponding tool service. When the Tool Capability receives a request for this new step and cannot find a suitable tool, the Workflow Capability informs the Tool Provider to register a new tool. After the Tool Provider registers a weather querying service with the Tool Broker (Figure 3 (3)), the user re-submits the same query. Consequently, when the workflow reaches the fourth step, the Tool Capability can now match and return information about the newly added tool service.

\section{Conclusion}

We propose the CACA Agent (Capability Collaboration based AI Agent), which introduces a set of capabilities working collaboratively to implement AI Agents. Drawing inspiration from service computing, the CACA Agent adopts an open architecture, effectively enhancing the extensibility of both planning abilities and tools for AI agents. Using the proposed system, we present a demonstration to illustrate the operational mechanism and application scenario extension mechanism of the CACA Agent.

Because LLMs in our design are utilized for domain-specific reasoning, the scale of the model could be effectively reduced while maintaining high inference quality\cite{luo2023taiyi}. We are currently transitioning the LLM that the Planning and Tool Capabilities rely on from LLM services (GPT-3.5 Turbo) to an LLM that can be deployed in a CPU environment. This transition aims to further increase the practicality and flexibility of AI Agents.

\bibliographystyle{ACM-Reference-Format}
\bibliography{sample-base}


\begin{thebibliography}{10}


\ifx \showCODEN    \undefined \def \showCODEN     #1{\unskip}     \fi
\ifx \showDOI      \undefined \def \showDOI       #1{#1}\fi
\ifx \showISBNx    \undefined \def \showISBNx     #1{\unskip}     \fi
\ifx \showISBNxiii \undefined \def \showISBNxiii  #1{\unskip}     \fi
\ifx \showISSN     \undefined \def \showISSN      #1{\unskip}     \fi
\ifx \showLCCN     \undefined \def \showLCCN      #1{\unskip}     \fi
\ifx \shownote     \undefined \def \shownote      #1{#1}          \fi
\ifx \showarticletitle \undefined \def \showarticletitle #1{#1}   \fi
\ifx \showURL      \undefined \def \showURL       {\relax}        \fi
\providecommand\bibfield[2]{#2}
\providecommand\bibinfo[2]{#2}
\providecommand\natexlab[1]{#1}
\providecommand\showeprint[2][]{arXiv:#2}

\bibitem[Huang et~al\mbox{.}(2022)]%
        {huang2022inner}
\bibfield{author}{\bibinfo{person}{Wenlong Huang}, \bibinfo{person}{Fei Xia}, \bibinfo{person}{Ted Xiao}, \bibinfo{person}{Harris Chan}, \bibinfo{person}{Jacky Liang}, \bibinfo{person}{Pete Florence}, \bibinfo{person}{Andy Zeng}, \bibinfo{person}{Jonathan Tompson}, \bibinfo{person}{Igor Mordatch}, \bibinfo{person}{Yevgen Chebotar}, {et~al\mbox{.}}} \bibinfo{year}{2022}\natexlab{}.
\newblock \showarticletitle{Inner monologue: Embodied reasoning through planning with language models}.
\newblock \bibinfo{journal}{\emph{arXiv preprint arXiv:2207.05608}} (\bibinfo{year}{2022}).
\newblock


\bibitem[Luo et~al\mbox{.}(2023)]%
        {luo2023taiyi}
\bibfield{author}{\bibinfo{person}{Ling Luo}, \bibinfo{person}{Jinzhong Ning}, \bibinfo{person}{Yingwen Zhao}, \bibinfo{person}{Zhijun Wang}, \bibinfo{person}{Zeyuan Ding}, \bibinfo{person}{Peng Chen}, \bibinfo{person}{Weiru Fu}, \bibinfo{person}{Qinyu Han}, \bibinfo{person}{Guangtao Xu}, \bibinfo{person}{Yunzhi Qiu}, {et~al\mbox{.}}} \bibinfo{year}{2023}\natexlab{}.
\newblock \showarticletitle{Taiyi: A bilingual fine-tuned large language model for diverse biomedical tasks}.
\newblock \bibinfo{journal}{\emph{arXiv preprint arXiv:2311.11608}} (\bibinfo{year}{2023}).
\newblock


\bibitem[Patil et~al\mbox{.}(2023)]%
        {patil2023gorilla}
\bibfield{author}{\bibinfo{person}{Shishir~G Patil}, \bibinfo{person}{Tianjun Zhang}, \bibinfo{person}{Xin Wang}, {and} \bibinfo{person}{Joseph~E Gonzalez}.} \bibinfo{year}{2023}\natexlab{}.
\newblock \showarticletitle{Gorilla: Large language model connected with massive apis}.
\newblock \bibinfo{journal}{\emph{arXiv preprint arXiv:2305.15334}} (\bibinfo{year}{2023}).
\newblock


\bibitem[Qin et~al\mbox{.}(2023)]%
        {qin2023toolllm}
\bibfield{author}{\bibinfo{person}{Yujia Qin}, \bibinfo{person}{Shihao Liang}, \bibinfo{person}{Yining Ye}, \bibinfo{person}{Kunlun Zhu}, \bibinfo{person}{Lan Yan}, \bibinfo{person}{Yaxi Lu}, \bibinfo{person}{Yankai Lin}, \bibinfo{person}{Xin Cong}, \bibinfo{person}{Xiangru Tang}, \bibinfo{person}{Bill Qian}, {et~al\mbox{.}}} \bibinfo{year}{2023}\natexlab{}.
\newblock \showarticletitle{Toolllm: Facilitating large language models to master 16000+ real-world apis}.
\newblock \bibinfo{journal}{\emph{arXiv preprint arXiv:2307.16789}} (\bibinfo{year}{2023}).
\newblock


\bibitem[Schick et~al\mbox{.}(2023)]%
        {schick2023toolformer}
\bibfield{author}{\bibinfo{person}{Timo Schick}, \bibinfo{person}{Jane Dwivedi-Yu}, \bibinfo{person}{Roberto Dess{\`\i}}, \bibinfo{person}{Roberta Raileanu}, \bibinfo{person}{Maria Lomeli}, \bibinfo{person}{Luke Zettlemoyer}, \bibinfo{person}{Nicola Cancedda}, {and} \bibinfo{person}{Thomas Scialom}.} \bibinfo{year}{2023}\natexlab{}.
\newblock \showarticletitle{Toolformer: Language models can teach themselves to use tools}.
\newblock \bibinfo{journal}{\emph{arXiv preprint arXiv:2302.04761}} (\bibinfo{year}{2023}).
\newblock


\bibitem[Song et~al\mbox{.}(2023)]%
        {song2023restgpt}
\bibfield{author}{\bibinfo{person}{Yifan Song}, \bibinfo{person}{Weimin Xiong}, \bibinfo{person}{Dawei Zhu}, \bibinfo{person}{Cheng Li}, \bibinfo{person}{Ke Wang}, \bibinfo{person}{Ye Tian}, {and} \bibinfo{person}{Sujian Li}.} \bibinfo{year}{2023}\natexlab{}.
\newblock \showarticletitle{RestGPT: Connecting Large Language Models with Real-World Applications via RESTful APIs}.
\newblock \bibinfo{journal}{\emph{arXiv preprint arXiv:2306.06624}} (\bibinfo{year}{2023}).
\newblock


\bibitem[Wang et~al\mbox{.}(2023)]%
        {wang2023survey}
\bibfield{author}{\bibinfo{person}{Lei Wang}, \bibinfo{person}{Chen Ma}, \bibinfo{person}{Xueyang Feng}, \bibinfo{person}{Zeyu Zhang}, \bibinfo{person}{Hao Yang}, \bibinfo{person}{Jingsen Zhang}, \bibinfo{person}{Zhiyuan Chen}, \bibinfo{person}{Jiakai Tang}, \bibinfo{person}{Xu Chen}, \bibinfo{person}{Yankai Lin}, {et~al\mbox{.}}} \bibinfo{year}{2023}\natexlab{}.
\newblock \showarticletitle{A survey on large language model based autonomous agents}.
\newblock \bibinfo{journal}{\emph{arXiv preprint arXiv:2308.11432}} (\bibinfo{year}{2023}).
\newblock


\bibitem[Wei et~al\mbox{.}(2022)]%
        {2022Chain}
\bibfield{author}{\bibinfo{person}{Jason Wei}, \bibinfo{person}{Xuezhi Wang}, \bibinfo{person}{Dale Schuurmans}, \bibinfo{person}{Maarten Bosma}, \bibinfo{person}{Ed Chi}, \bibinfo{person}{Quoc Le}, {and} \bibinfo{person}{Denny Zhou}.} \bibinfo{year}{2022}\natexlab{}.
\newblock \showarticletitle{Chain of Thought Prompting Elicits Reasoning in Large Language Models}.
\newblock  (\bibinfo{year}{2022}).
\newblock


\bibitem[Weng(2023)]%
        {weng2023prompt}
\bibfield{author}{\bibinfo{person}{Lilian Weng}.} \bibinfo{year}{2023}\natexlab{}.
\newblock \showarticletitle{LLM-powered Autonomous Agents}.
\newblock \bibinfo{journal}{\emph{lilianweng.github.io}} (\bibinfo{date}{Jun} \bibinfo{year}{2023}).
\newblock
\urldef\tempurl%
\url{https://lilianweng.github.io/posts/2023-06-23-agent/}
\showURL{%
\tempurl}


\bibitem[Yao et~al\mbox{.}(2022)]%
        {yao2022react}
\bibfield{author}{\bibinfo{person}{Shunyu Yao}, \bibinfo{person}{Jeffrey Zhao}, \bibinfo{person}{Dian Yu}, \bibinfo{person}{Nan Du}, \bibinfo{person}{Izhak Shafran}, \bibinfo{person}{Karthik Narasimhan}, {and} \bibinfo{person}{Yuan Cao}.} \bibinfo{year}{2022}\natexlab{}.
\newblock \showarticletitle{React: Synergizing reasoning and acting in language models}.
\newblock \bibinfo{journal}{\emph{arXiv preprint arXiv:2210.03629}} (\bibinfo{year}{2022}).
\newblock


\end{thebibliography}





\end{document}